\documentclass[preprint,review,12pt]{elsarticle}

\usepackage{amssymb}
\usepackage[fleqn]{amsmath}
\usepackage{multirow}

\usepackage{float}
\usepackage{graphicx}
\usepackage{amsfonts,mathtools}
\newcommand*{\underuparrow}[1]
{\ensuremath{\underset{#1}{\uparrow}}}
\newcommand*{\underdownarrow}[1]
{\ensuremath{\underset{#1}{\downarrow}}}

\newcommand*{\underequiv}[1]
{\ensuremath{\underset{#1}{\equiv}}}
\newcommand*{\undernsim}[1]
{\ensuremath{\underset{#1}{\nsim}}}

\usepackage{algorithm}
\usepackage{algorithmicx}
\usepackage{algpseudocode}
\usepackage{caption}
\usepackage{subcaption}

\usepackage{hyperref}
\usepackage[dvipsnames]{xcolor}

\usepackage{tcolorbox}
\tcbuselibrary{theorems}
\newtcbtheorem[number within=section]{example}{Example}%
{colback=gray!5,colframe=black,fonttitle=\bfseries}{exa}
\newtcbtheorem[number within=section]{definitionSC}{Definition}%
{colback=gray!5,colframe=cyan!90!green!40,fonttitle=\bfseries,coltitle=black}{def}
\newtcbtheorem[use counter from=definitionSC, number within=section]{definitionSI}{Definition}%
{colback=gray!5,colframe=pink!50!orange,fonttitle=\bfseries,coltitle=black}{def}
\newtcbtheorem[use counter from=definitionSI, number within=section]{definitionG1}{Definition}%
{colback=gray!5,colframe=blue!60!green!50!black!90,fonttitle=\bfseries}{def}
\newtcbtheorem[use counter from=definitionG1,number within=section]{definitionG2}{Definition}%
{colback=gray!5,colframe=red!50!black!65!orange,fonttitle=\bfseries}{def}
\newtcbtheorem[use counter from=definitionG2,number within=section]{definitionG3}{Definition}%
{colback=gray!5,colframe=yellow!80!red!80!green!80,fonttitle=\bfseries,coltitle=black}{def}

\journal{Knowledge-Based Systems}

\begin{document}

\begin{frontmatter}

\title{Semantic Abstraction for Natural Language Inference: a Methodological Framework for Discovering and Compensating Semantic Knowledge and Reasoning Gaps in Large Language Models}
\nonumnote{This is a preprint of an article published in Knowledge-Based Systems. The final authenticated version is available online at: \url{https://doi.org/10.1016/j.knosys.2025.114825}}

\author[cinc]{David Torres-Moreno
\fnref{conacyt}
}
\author[cinc]{Jorge Hermosillo-Valadez\corref{corresp}}%
\cortext[corresp]{Corresponding author.}\ead{jhermosillo@uaem.mx}
\fntext[conacyt]{The author would like to thank CONACYT MEXICO for the grant awarded to carry out this work.}

\affiliation[cinc]{organization={Centro de Investigación en Ciencias, Universidad Autónoma del Estado de Morelos},%
            addressline={Av. Universidad 1001},
            city={Cuernavaca},
            postcode={62209},
            state={Morelos},
            country={México}}

\begin{abstract}
Despite their outstanding performance on many NLP tasks, LLMs face serious challenges related to semantic abstraction.  In this study, we are interested in understanding how LLMs leverage abstract semantic knowledge in natural language inference (NLI), which requires sophisticated linguistic capabilities to interpret implicit meanings, contextual conceptual relationships, and semantic connections between words and phrases. To this end, we propose a methodological framework for constructing new semantic knowledge at a higher level of abstraction, which we define under the notions of semantic compatibility and incompatibility for NLI. In this framework, the meaning of the lexical-semantic relations between the premise and the hypothesis is reconfigured to achieve a more flexible semantic network that induces different reasoning paths in LLMs. These new pathways show a consistent pattern of responses that allows agreement on a single response. The results demonstrate that our proposal allows to discover and compensate for LLMs' semantic knowledge gaps in NLI, achieving significant improvements in accuracy, exceeding 10\% for some models, and in particular for the non-entailment class. It is essential to note that LLMs need structured knowledge and not just more data to bridge reasoning gaps. Our hybrid approach directs attention to overlooked word relationships, allowing models to synthesize missing information. We believe that the future lies not in increasing model size, but in creating a semantic scaffolding that mimics the flexibility of human thinking. Hopefully, our proposal will enable the development of more robust agents and interpretable reasoning, guiding AI toward reliable language understanding.
\end{abstract}

\begin{graphicalabstract}
\centering%
\includegraphics[width=\linewidth]{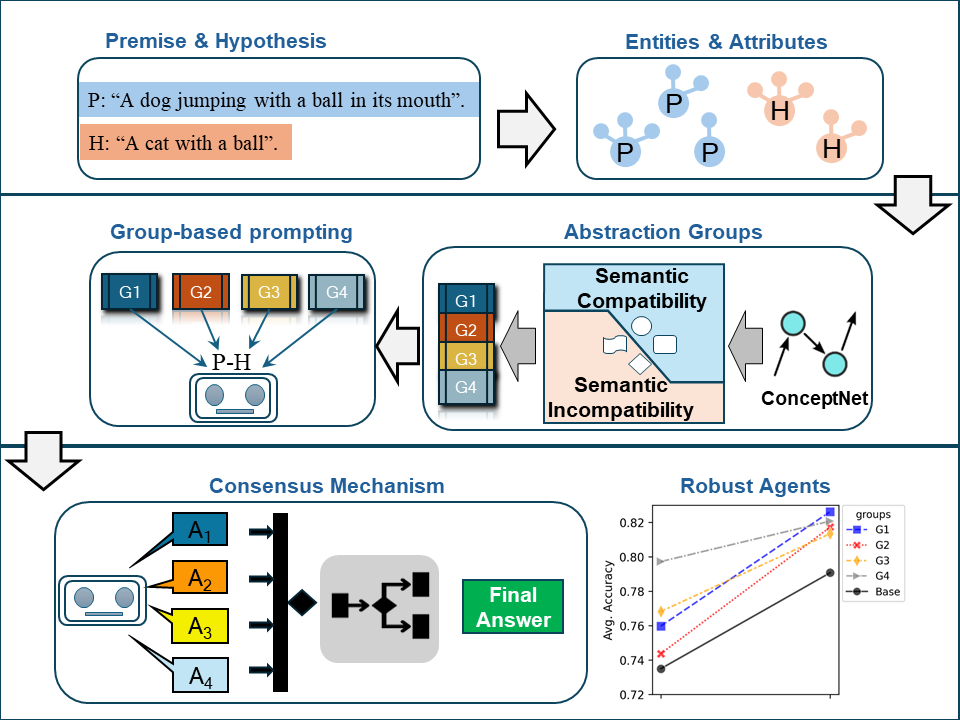}
\end{graphicalabstract}

\begin{highlights}
\item A methodological framework to discover and compensate for semantic knowledge gaps of Large Language Models (LLMs) in Natural Language Inference (NLI).
\item The method guides the LLMs' reasoning process, revealing their weaknesses and inconsistencies, which are compensated for by explainable decision strategies that improve NLI accuracy by up to 10\% for some models.
\item The results show that the framework is useful for improving not-entailment inference and suggest that it could also contribute to develop more robust and reliable models in the field of Natural Language Understanding (NLU).

\end{highlights}

\begin{keyword}
LLMs
\sep NLI
\sep Textual Entailment
\sep Abstraction
\sep Semantic Knowledge

\end{keyword}

\end{frontmatter}

\section{Introduction}
\label{sec1}
\subsection{Large language models and Natural Language Inference}

Large language models (LLMs) have been a milestone in the history of artificial intelligence, having demonstrated their degree of capability in \textit{understanding, knowledge, reasoning and calculation}
\citep{zhong-etal-2024-agieval,  zhao2025surveylargelanguagemodels}.
Despite an outstanding performance in many Natural Language Processing (NLP) tasks, LLMs exhibit critical problems.
It has been shown, for instance, that poor knowledge or obvious factual gaps in training data have an impact on the phenomenon of hallucinations in LLMs \citep{manakul-etal-2023-selfcheckgpt,rawte-etal-2023-troubling,dhuliawala-etal-2024-chain}. Moreover, if the training process is contaminated with task evaluation data a mirage of competence is created, evidenced by models that memorize specific examples but fail at new tasks \citep{mccoy-etal-2019,jin2020,Li_Flanigan_2024}, or reproduce
problems inherited from human data, such as social biases and annotation artifacts
 \citep{proebsting-poliak-2025-biases}. To compensate knowledge gaps, it has been shown that providing external knowledge graphs can improve the quality of embedded representations \citep{YANG2022109298}, mitigate hallucination \citep{li-etal-2023-halueval}, and helps answer fact-intensive questions better \citep{DAI2025113060}.

On the other hand, even more worrying is their inability to reason in depth. The inability of these models to perform abstract reasoning has been highlighted \citep{NEURIPS2024_da5498f8,wang-etal-2024-llms,he-etal-2025-language}. While there are hints that pre-trained language models (PLMs) recognize differences between hyponym-hypernym relationships in nouns \citep{regneri-etal-2024-detecting}, higher-level cognitive limitations have been observed  \citep{peng-etal-2022-copen}. Furthermore, LLMs lack functional and formal linguistic skills \citep{MAHOWALD2024517}, and show fundamental limitations in abstract reasoning and planning \citep{Lee_2025_abstraction}.

In this study, we are interested in understanding how LLMs leverage abstract semantic knowledge and to what extent this is useful to their reasoning process and linguistic skills. To assess the usefulness of abstraction we turn to Natural Language Inference (NLI), which requires sophisticated linguistic capabilities to interpret implicit meanings, contextual concept relationships, and semantic connections between words and phrases. Addressing NLI effectively is a matter of great relevance, since it is the basis for other NLP tasks with potential applications in healthcare \citep{jullien-etal-2023-semeval, FEI2021106962}, legal domain \citep{wehnert2022} or fact checking \citep{MARTIN2022109265, thorne-etal-2018-fever}.

NLI examines whether a Hypothesis ($H$) can be inferred from a given Premise ($P$) \citep{dagan2006}. The problem is to decide whether to assign an \textit{Entailment} label when $P\to H$ can be evaluated to $true$, a \textit{Contradiction} label when $P\to H$ is considered $false$, or a \textit{Neutral} tag when $P\to H$ is $undetermined$. Alternatively,  the recognition of textual entailment (RTE) task considers two target classes: \textit{Entailment} and \textit{Not-entailment}.

LLMs can effectively address NLI when trained on benchmark datasets \citep{madaan2024lostinferencerediscoveringrole}, but stumble in scenarios that require abstract thinking, revealing a worrying dependence on superficial shortcuts that becomes evident in cognitive robustness tests  \citep{huber-niklaus-2025-llms,gururangan-etal-2018-annotation}.
Moreover, LLMs show poor cognitive flexibility, as they do not capture the nuances of human disagreement \citep{lee-etal-2023-large, nie-etal-2020-learn} and show sensitivity to domain shift \citep{sadat-caragea-2024-mscinli}. This has also been  underscored by \citep{WANG2025100103}, who shows that the conceptual networks of LLMs are less interconnected and less flexible than those of humans, which manifests itself in more sequential reasoning, difficulty in jumping between distant semantic domains, and limitations in analogical reasoning. This cognitive rigidity suggests a fundamentally different structure of semantic relationships than in humans and points to the need to develop models with dynamic semantic networks to achieve advanced reasoning.

\subsection{Problem statement and contribution}

The previous analysis elucidates the broad limitations
and gaps in LLMs' reasoning capabilities and semantic understanding for NLI tasks: \emph{LLMs struggle with abstract reasoning, relying on superficial cues and rigid, domain-limited conceptual networks that hinder cognitive flexibility}.

We propose a methodological framework for abstracting semantic knowledge in a hierarchical way that helps to identify and fill knowledge gaps of LLMs regarding NLI.  The key of our proposal is to model cognitive aspects that are implicit in NLP \citep{hupkes2023_taxo}. We propose that in NLI the meaning of the Hypothesis should be contained in the Premise, but \emph{the semantic content of H is usually more general than that of P when the entailment relation is true}. This can be justified by observing that humans can identify semantic relationships at different levels of abstraction that allow us to make entailment decisions.
Consider the following example, where the problem is to decide whether $P\to H1$, or $P\to H2$ is true ---for clarity, we present the idea in the context of RTE:
\begin{itemize}
    \item[P:] ``A dog jumping with a ball in its mouth".
    \item[H1:] ``An animal with a ball".
    \item[H2:] ``A cat with a ball".
\end{itemize}
In our approach, it is not just a matter of finding contextual similarities, but to know the multiple semantic relations that contribute to the decision of the fact that $P$ entails $H$.
In this case, the use of external resources of knowledge such as ConceptNet \citep{Speer2016ConceptNet5A} is of great help.

A trivial approach would be to navigate the network of ConceptNet so as to find out if there is a connection between two concepts; for example, ``dog \emph{is a type of} mammal" and ``mammal \emph{is a type of} animal" are actual pathways in ConceptNet. Eventually, we could establish a link between ``dog" and ``animal", but we could also establish a link between ``dog" and ``cat" using the same navigation scheme.

However, under this approach there is no notion of hierarchy between concepts: \emph{the navigation does not allow us to know whether ``dog" is more general than ``cat" or whether they are at the same level}; we cannot establish semantic links of another level of abstraction. Take another example, ``Stockholm \emph{is part of} Sweden" and ``Sweden \emph{is a type of} Europe" are also pathways in ConceptNet that allow to build the direct relation ``Stockholm \emph{is part of} Europe", similar to the \emph{is a} relation.
The semantic similarity between ``is part of", ``is a type of" and ``is a" allows to abstract a notion of \emph{generality}: ``animal" is more general than ``dog", ``Europe" is more general than ``Stockholm", but ``dog" is not more general than ``cat" because there is another relation between them that is ``distinct from".

This notion of generality is what we exploit to establish links between P and H. Thus, this abstraction of generality must exist in only one direction from P to H to infer that entailment is true. Back to our example above, $P\to H1$ is $true$ and $P\to H2$ is $false$. In the first case, we know the relationship ``dog is an animal" and the asymmetric relationship between ``dog" (hyponym) and ``animal" (hyperonym) hold ---hyperonymy conveys the notion of a more general concept. But, in the case $P\to H2$, the relationship ``dog is distinct from cat" and the co-hyponym relationship between ``dog" and ``cat" also hold ---co-hyponym conveys the notion that two concepts share the same hyperonym. The former relations directly support entailment, whereas the latter do not.

We argue that it is not necessary to know the details (definition) of a specific type of relationship, but rather that it is sufficient to abstract these relationships into categories that entail semantic compatibility.
Our hypothesis is that either LLMs do not know which relation is useful, or they ignore how to abstract them into semantically compatible categories allowing it to decide for a particular class of entailment.

We propose a new methodological framework of abstraction of semantic relations that allows to improve the reasoning process of an LLM and to know to what extent these abstractions are useful in NLI.
In this way, we not only inject new knowledge to the model but also provide new semantic connections between concepts, making the conceptual network of the model more flexible. Our contribution is as follows:

\begin{itemize}
    \item We propose a methodological framework to discover and compensate for semantic knowledge gaps of LLMs in NLI.
    \item The method considers the directionality required in entailment evaluation to define abstract classes of semantic compatibility and incompatibility that guide the LLMs' reasoning process, revealing inconsistencies in responses and weaknesses.
    \item The framework provides flexibility to the reasoning and decision pipeline in that semantic compatibility and incompatibility relations allow for new valid connections between concepts, distinct inference paths and the implementation of robust decision strategies.
    \item The results show that the framework is particularly useful for improving not\_entailment inference and suggest that it could also contribute to develop more robust and reliable agents in the field of NLU.

\end{itemize}

The remainder of the paper is structured as follows: Section \ref{sec:relwork} discusses the related work; Section \ref{sec:method} introduces the detailed methodological framework; Section \ref{sec:exp} presents the experimental setup and results; Section \ref{sec:disc} discusses the implications of the findings and limitations of the work; and Section \ref{sec:concl} concludes with key contributions and final thoughts.

\section{Related work.}\label{sec:relwork}

While humans can combine simple concepts to solve complex problems or generate novel solutions, the performance of artificial neural models, such as PLMs and LLMs, deteriorates rapidly when faced with tasks that require abstraction.
Despite advances in versatile generation and reasoning, these models lack deep understanding of abstract concepts and exhibit systematic failures in semantic organization \citep{peng-etal-2022-copen}, generalization of dimensions \citep{dutt-etal-2024-investigating} and the identification  of abstract semantics and semantic relations \citep{beloucif-biemann-2021-probing-pre,rozanova-etal-2022-decomposing,cao2024CES}.
For instance, the authors in \citep{wang-etal-2024-abspyramid} analyze the dependence of their ability to critically handle abstract concepts.
Their results suggest that abstract thinking does not arise spontaneously, and that LLMs lack consistency in their inferential processes. Another example is \citep{Lee_2025_abstraction},
who show that these models often arrive at correct answers based on superficial patterns without understanding the underlying principles.

 This limitation
 may be related to an inability to maintain stable symbolic representations across varying contexts \citep{al-saeedi-harma-2025-emergence}. Interestingly, studies on attention maps in BERT reveal that PLMs can learn abstract relationships (such as hypernymy), but only implicitly and in a way that is biased by lexical distribution \citep{regneri-etal-2024-detecting}. This creates a problem: while these models capture certain hierarchical structures, they do not use them as flexibly or consistently as humans do.

The observed limitations demonstrate models' current inability to fully capture the nuanced reasoning, abstraction and linguistic sophistication required in NLI tasks.
To effectively tackle NLI, we can find work that seeks to provide new semantic information, either through new lexical-syntactic structures or through external resources. Thus, \citep{KIM2024112608} introduce virtual links between premises and hypotheses in order to overcome limitations based on lexical heuristics by extending syntactic structures, so that the model focuses on deeper contextual understanding.
On the other hand, another common strategy is to incorporate structured external resources that provide ontological and semantic information to mitigate hallucinations in their responses \citep{agrawal-etal-2024-knowledge}. External knowledge improves the performance of textual entailment through various approaches: knowledge graph-text fusion \citep{Wang2018ImprovingNL, Kapanipathi_2020}, dynamic semantic integration \citep{Guo2022}, and adapter-based knowledge incorporation \citep{lauscher-etal-2020-common}.

In order to cover the greatest number of semantic relationships, various knowledge bases can be used, which means dealing with a wide variety of representation approaches. The heterogeneity of available knowledge resources has led \citep{ILIEVSKI2021107347} to develop a unified dimensional framework for relationship classification. However, this presents fundamental challenges, particularly in ambiguous dimensional mapping, where relationships often transcend single categories and the optimal dimension for a task may not be the most useful. This complexity highlights the urgency of developing innovative methods that requires a delicate balance between semantic flexibility and structural precision to address abstract reasoning problems.

All this suggest that the challenge lies not only in scale, but also in cognitive architecture. While the current approach of predicting tokens allows some abstraction to emerge, it seems inadequate for replicating the compositional and abstractive mechanisms of human thought. The true test will be whether they can transcend statistical imitation and achieve a genuine structural understanding. Even with advanced methods such as Abstraction-of-Thought (AoT) \citep{hong-etal-2024-abstraction}, which forces hierarchical reasoning (from the abstract to the concrete), LLMs still demonstrate limitations in tasks requiring deep generalization, as they have an interpretative rigidity \citep{sedova-etal-2024-know} and a cognitive bias \citep{echterhoff-etal-2024-cognitive}.

Therefore, we provide a methodological framework for abstracting semantic relationships between the premise and the hypothesis to a higher semantic level, so that models can grasp the missing generalities to effectively tackle the NLI task.

Our proposal not only identifies current limitations but also establishes conceptual bridges between available explicit knowledge and the implicit reasoning processes underlying the LLMs' linguistic understanding.
We propose a categorical abstraction framework through the structured extraction of semantic relations from ConceptNet to build semantic abstraction groups that are used to prompt models. Our approach goes beyond the mere retrieval of semantic relations; its goal is to build new semantic knowledge at a higher level of abstraction, where the meaning of relations is given by their task-oriented association.
We seek to make this new knowledge work to induce reasoning paths in LLMs, different from those they would follow without this new knowledge, in a direction consistent with the classes of NLI tasks. Thus, we investigate how our categorization allows us to discover and compensate for semantic knowledge gaps in LLMs in NLI.

\section{Methods}\label{sec:method}
Inspired by the way humans solve NLI tasks, we have developed a framework that incorporates key cognitive processes, such as meaning abstraction, semantic interpretation, and inference.

\begin{figure}[h]%
\centering%
\includegraphics[width=\linewidth]{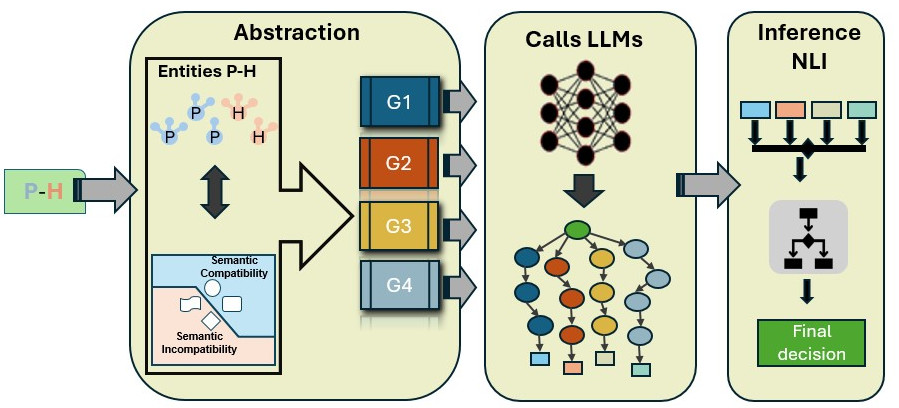}
\caption{Methodological framework}\label{fig:fig2_Met_framework}
\end{figure}

The framework is summarized in Figure \ref{fig:fig2_Met_framework}. We start by defining two abstract categories: \textbf{Semantic Compatibility} and \textbf{Semantic Incompatibility}. The abstraction step involves the analysis of the dependency tree of $P$ and $H$ to identify entities---nouns and verbs---and their corresponding attributes---modifiers such as adjectives and adverbs. We use ConceptNet to identify the semantic relationships between these entities and group them according to the definition of the abstract categories. These groups of semantic relations constitute the formal mechanism used to prompt the LLMs. From LLMs' answers it is possible to analyze the reasoning paths and eventually propose a decision model.%

We now describe in detail the methodological framework.

\subsection{Semantic Compatibility and Incompatibility}
Consider two concepts $c_1$ and $c_2$. We define semantic compatibility between them as follows:

\begin{definitionSC}{Semantic Compatibility}{sc}
We say that $c_2$ has a relation of Semantic Compatibility with $c_1$ if there is a relation from $c_1$ to $c_2$
and $c_2$ is at a \textbf{higher} level in the hierarchy of the conceptual network (\textbf{generalization}) or $c_2$ is at the same level with an \textbf{equivalent} meaning (\textbf{equivalence}). We write
$c_1 \underuparrow{} c_2$ for the generalization relation and $c_1 \underequiv{} c_2$ for the equivalence relation.
\end{definitionSC}

Semantic compatibility helps to preserve meaning. For example, if we ask whether ``person" has a relation of semantic compatibility with ``man", or if $man\underuparrow{}person$ holds, the answer is yes, since ``person" is hyperonym of ``man"---``person" is at a higher level in the hierarchy of the conceptual network; it generalizes ``man". Thus, the semantic compatibility relationship operates under a strict directional constraint. The inverse relation $person\underuparrow{} man$ fails to preserve the meaning, since ``person” could refer to a ``woman". This is an example of semantic incompatibility that we define below.

\begin{definitionSI}{Semantic Incompatibility}{si}
We say that $c_2$ has a relation of Semantic Incompatibility with $c_1$ if there is a relation from $c_1$ to $c_2$ and $c_2$ is at a \textbf{lower} level in the hierarchy of the conceptual network (\textbf{concretization relation}), we write $c_1 \underdownarrow{} c_2$, or if there is a relation from $c_1$ to $c_2$ and $c_2$ is at the same level with a \textbf{different} meaning (\textbf{opposition relation}), we write $c_1 \undernsim{} c_2$.
\end{definitionSI}

Semantic incompatibility helps to identify relations that either concretize concepts or create a direct contrast (or difference in meaning) between concepts. On the one hand, recalling the example above, ``person" is concretized in ``man", i.e. the relation $person\underdownarrow{} man$ holds. On the other hand, the most obvious cases of opposition are antonyms, but also more subtle differences in meaning are derived from co-hyponymy where terms share a hyperonym but are mutually exclusive. For example ``running" and ``walking" are co-hyponyms under their shared hyperonym ``movement". Thus, we say that ``running" has a relation of semantic incompatibility with ``walking"; i.e. $running\undernsim{} walking$ holds.

Notice that the Semantic Compatibility relation is reflexive, symmetric for $c_1 \underequiv{} c_2$,  and antisymmetric and transitive for $c_1 \underuparrow{} c_2$. These properties can be observed in different types of semantic relations in conceptual networks. Reflexivity occurs when a concept is related to itself, while asymmetry appears in generalization or concretization relations. On the other hand, transitivity in generalization relations is bottom-up, enabling hierarchical inferences: if ``dog" is a ``mammal" and ``mammal" is an ``animal", then ``dog" is an ``animal" that preserve meaning according to the direction of the relation.

Notice also that the Semantic Incompatibility relation is symmetric for $c_1 \undernsim{} c_2$ and transitive for the concreteness relationship ($c_1 \underdownarrow{} c_2$). However, unlike the generalization relation, which is also transitive, the direction of concreteness relation is top-down and does not preserve meaning. Thus, these properties organize the conceptual network, facilitating attribute inheritance and logical reasoning within the conceptual structure.

\begin{figure}[h]
  \centering
  \includegraphics[width=0.7\linewidth]{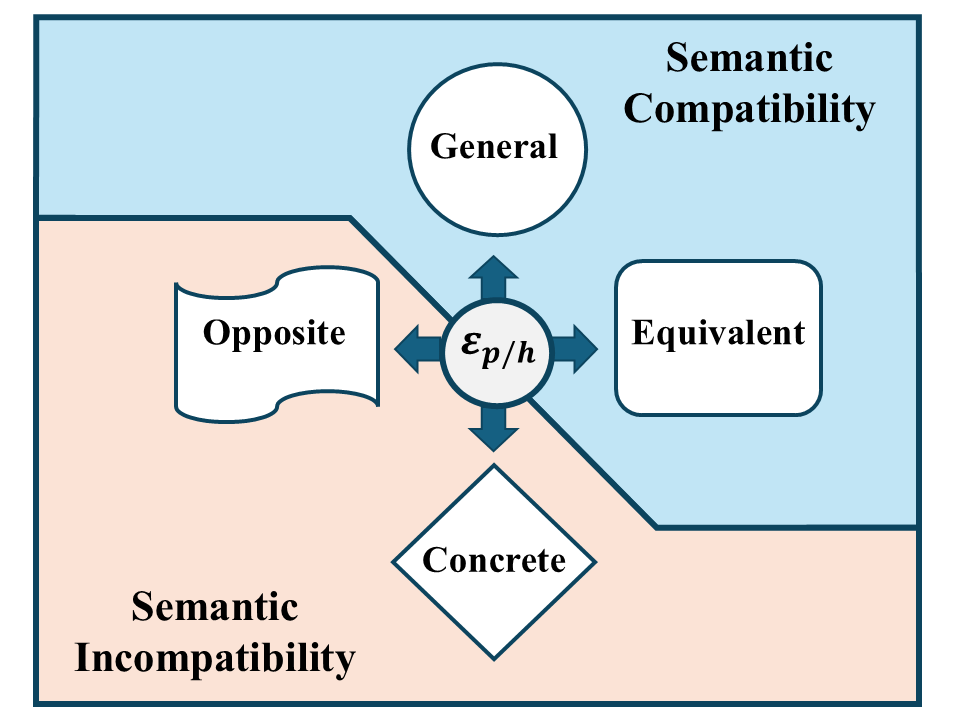}
  \caption{Concept abstraction framework as given by definitions \ref{def:sc} and \ref{def:si}.}\label{fig:fig3_CS5}
\end{figure}

These conceptual abstractions are crucial for semantic analysis, particularly when evaluating entailment. Our framework uses ConceptNet's predefined relationships. Hence, it is possible to identify the type, direction and interaction of connections between lexical units and sub-phrases. In the conceptual hierarchy, upward vertical links allow the establishment of logical entailment. These can be combined with horizontal links representing equivalence relations (synonyms) belonging to semantic compatibility. Horizontal links of opposite relationships (antonyms and co-hyponyms), and downward vertical links belong to semantic incompatibility rather indicating not\_entailment. These conceptual abstractions are summarized in Figure \ref{fig:fig3_CS5}.

In order to clearly establish the links between semantic relations and entailment classes, we propose to group the former as described in the following section.

\subsection{Semantic Relationship Groups}
We now propose rules for associating semantic relationships that are consistent with definitions \ref{def:sc} and \ref{def:si}. These rules allow the construction of groups that will eventually allow to establish a link between semantic relations and the notion of entailment. Thus, the definition of groups aims to establish comparison rules between entities with attributes of both $P$ and $H$ to determine what kind of semantic compatibility relation they have. Therefore, we establish the following notation for pairing entities with attributes for both P and H.

For each Premise $P$ and each Hypothesis $H$ we consider the sets $\mathcal{E}_p:=\{(e^p_{i}[;\alpha^p_{1},\ldots,\alpha^p_{k}])|i=1,\ldots,p_1;\,k=1,\ldots,p_2\}$, where $e^p_{i}$ is an entity of $P$ and $\alpha^p_{1},\ldots,\alpha^p_{k}$ is an optional list of corresponding attributes, and $\mathcal{E}_h:=\{(e^h_{i}[;\alpha^h_{1},\ldots,\alpha^h_{k}])|i=1,\ldots,h_1;\,k=0,1,\ldots,h_2\}$, where $e^h_{i}$ is an entity of $H$ and $\alpha^h_{1},\ldots,\alpha^h_{k}$ is an optional list of corresponding attributes. With this notation, each entity and attribute can be treated separately, thus, in order to recover the original phrase we will write $\varepsilon^*$ to represent an entity with its corresponding attributes in natural language. All entities and attributes are lemmatized. The following example shows the workings of this notation for given $P$ and $H$.

\begin{example}{Entity and attributes notation}{e1}
\vspace{-1em}
\begin{small}
$$
\begin{array}{ll}
     P: &  \text{An old man in a long-sleeves white shirt is walking to work} \\
     & \text{in a big city.}\\
     \mathcal{E}_p= & \{(man;old),(shirt;white,long\text{-}sleeve),(walk),(work),(city;big)\}\\
      \varepsilon_2= & (shirt;white,long\text{-}sleeve); \varepsilon_2^*=long\text{-}sleeve\;white\;shirt\vspace{1em}\\
    H: &  \text{The man is wearing shorts and a t-shirt as he jogs.}\\
    \mathcal{E}_h= & \{(man),(wear),(short),(t\text{-}shirt),(jog)\}
\end{array}
$$
\end{small}
\end{example}

\begin{definitionG1}{Group 1: General and Equivalent Relations}{g1}
Given two elements $\varepsilon_p\in\mathcal{E}_p$ and $\varepsilon_h\in\mathcal{E}_h$. We say that $\varepsilon_h$ has a relation of \textbf{Semantic Compatibility} with $\varepsilon_p$, if either $e^p\underuparrow{}e^h$ or $e^p\underequiv{}e^h$ hold and, whenever $\alpha_j^h$ exists, either $\alpha_k^p\underuparrow{}\alpha_j^h$ or $\alpha_k^p\underequiv{}\alpha_j^h$ hold for all $k,j$.
In this case,  we say that the relation belongs to \textbf{Group 1 ($G_1$)}. We write
$(\varepsilon_p^*, \text{rel\_SC},\varepsilon_h^*)\in G_1$, where rel\_SC is any valid relation in the concept graph.

\end{definitionG1}

\begin{definitionG2}{Group 2: Opposite and Difference Relations}{g2}
Given two elements $\varepsilon_p\in\mathcal{E}_p$ and $\varepsilon_h\in\mathcal{E}_h$. We say that $\varepsilon_h$ has a relation of \textbf{Semantic Incompatibility (Difference)} with $\varepsilon_p$, if either $e^p\undernsim{}e^h$ or $\alpha^p_k\undernsim{}\alpha^h_j$ hold for any $k,j$.
In this case, we say that the relation belongs to \textbf{Group 2 ($G_2$)}.  We write
$(\varepsilon_p^*, \text{rel\_SID},\varepsilon_h^*)\in G_2$, where rel\_SID is any valid relation in the concept graph.
\end{definitionG2}

\begin{definitionG3}{Group 3: Concrete Relations}{g3}
Given two elements $\varepsilon_p\in\mathcal{E}_p$ and $\varepsilon_h\in\mathcal{E}_h$, we say that $\varepsilon_h$ has a relation of \textbf{Semantic Incompatibility (Concrete)} with $\varepsilon^p$, if either $e^p \underdownarrow{} e^h$ or $\alpha^p_k \underdownarrow{} \alpha^h_j$ hold for any $k,j$. In this case, we say that the relation belongs to \textbf{Group 3 ($G_3$)}. We write
$(\varepsilon_p^*, \text{rel\_SIC},\varepsilon_h^*)\in G_3$, where rel\_SIC is any valid relation in the concept graph.
\end{definitionG3}

Note that the term ``valid” refers to a relationship in the knowledge graph that complies with the definitions of Semantic Compatibility \ref{def:sc} and \ref{def:si}. The last group $G_4$ will gather all entities of $H$ that do not have any relation with any of the entities of $P$. In that case, we write
$$(,\text{UNK},\varepsilon_h)\in G_4.$$

Continuing with Example \ref{exa:e1}, the following lists constitute the groups as follows, where the relations are extracted from ConceptNet:
\begin{example}{Groups of relations from Example \ref{exa:e1}}{e2}
\vspace{-1em}
\begin{small}
$$
\begin{array}{ll}
   G_1:=  &  \big[\big(old\;man,\text{\textbf{is a}},man\big),\big(long\text{-}sleeve\;white\;shirt,\text{\textbf{is a}},t\text{-}shirt\big)\big]\\
   G_2:=  &  \big[\big(walk,\text{\textbf{distinct from}},jog\big),\\
   & \;\;\big(long\text{-}sleeve\;white\;shirt,\text{\textbf{distinct from}},short\big)\big]\\
   G_3:=  &  \big[\big(long\text{-}sleeve\;white\;shirt,\text{\textbf{related to}},wear\big)\big]\\
   G_4:=&\emptyset
\end{array}
$$
\end{small}
\end{example}

For the sake of clarity in the exposition of our arguments, in what follows we will make an abuse of language and will refer to entities with attributes as just entities.

\subsection{Flexible conceptual networks: creating new connections between concepts}
\label{sec:flexibility}
\begin{figure}[h!]
\centering
\begin{subfigure}[b]{0.62\linewidth}
\includegraphics[width=0.9\linewidth]{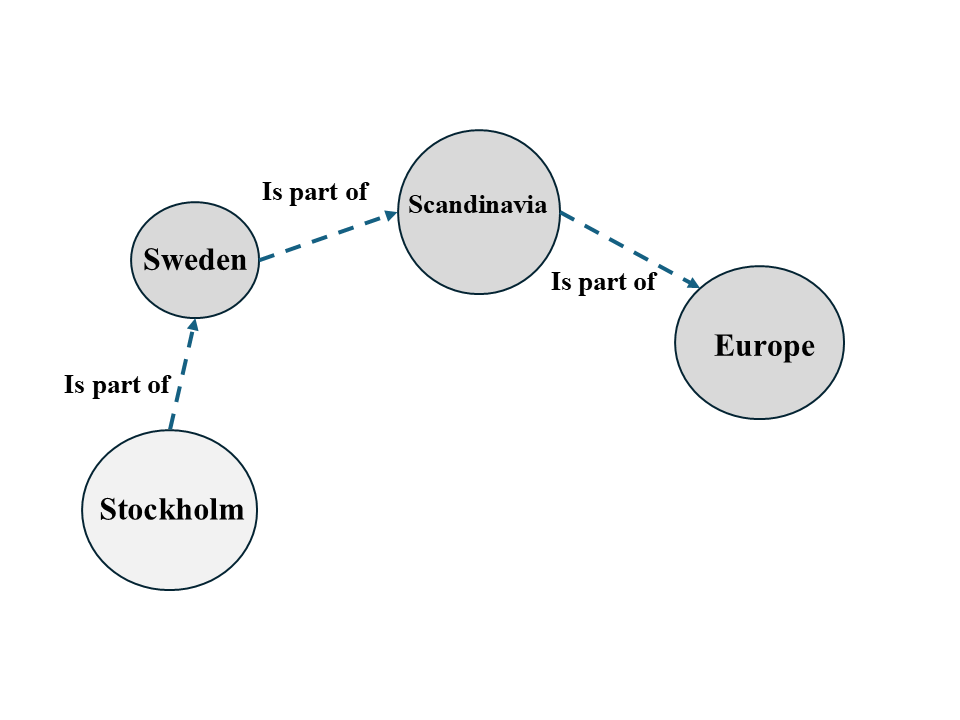}
\caption{}
\label{fig:path_concepts}
\end{subfigure}
\begin{subfigure}[b]{0.33\linewidth}
\includegraphics[width=\linewidth]{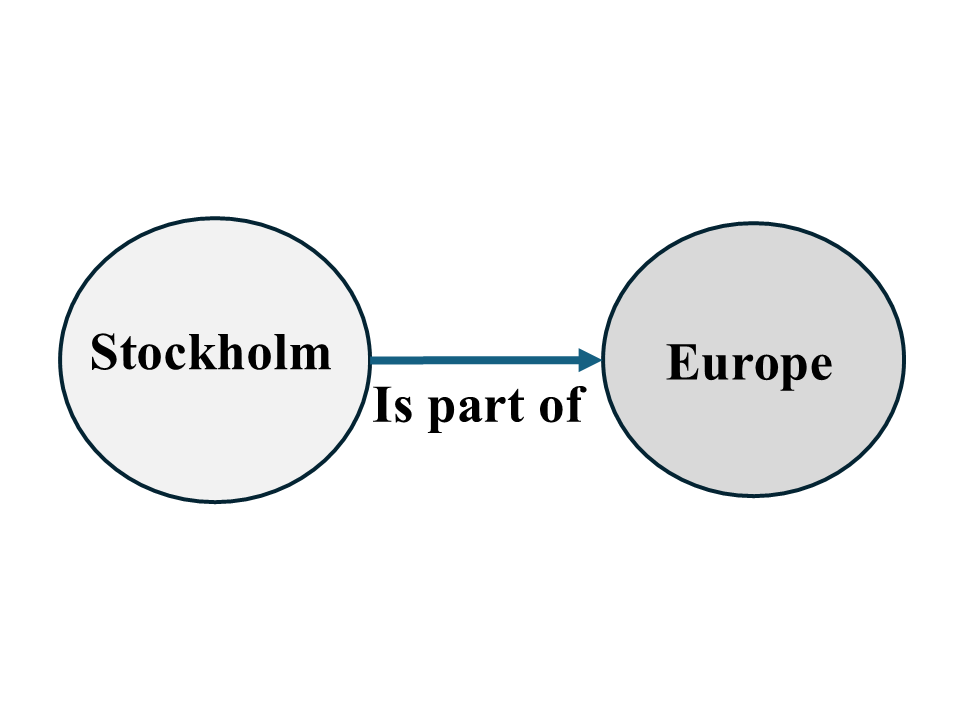}
\caption{}
\label{fig:newconcept}
\end{subfigure}
\caption{Flexibilization of the conceptual network through the construction of new connections between concepts. a) The generalization relation allows reaching the ``Europe" node from ``Stockholm", while maintaining Semantic Compatibility (with the \emph{is part of} relation) along the path. b) The transitivity of the rule allows to establish the direct link between these two concepts.}
\label{fig:fig4_groups}
\end{figure}

Previous groups define lists of nodes---relationships---or subgraphs connecting directly entities of $P$ and $H$.
Since we are going to navigate in a knowledge graph, we may or may not find direct relationships between concepts. In the first case, the relationship is established in a trivial way and can be readily categorized using the defined rules. If not found, the proposal is to try to establish a link between entities of P and H by identifying possible intersections between their subgraphs, enabling precise and consistent Semantic Compatibility scope. Figure \ref{fig:fig4_groups} illustrates this purpose using the example given in the Introduction about ``Stockholm".

In order to enrich the groups with non-direct semantic relations between entities of P and H, we propose to follow the guidelines of the Semantic Compatibility (SC) and Incompatibility (SI) rules using the transitivity property. That is, we want to know if there are other concepts having SC relations with entities of H that in turn connect with entities of P under this same principle.

\begin{figure}[h!]
\centering
\begin{subfigure}[b]{0.9\linewidth}
\centering
\includegraphics[width=0.9\linewidth]{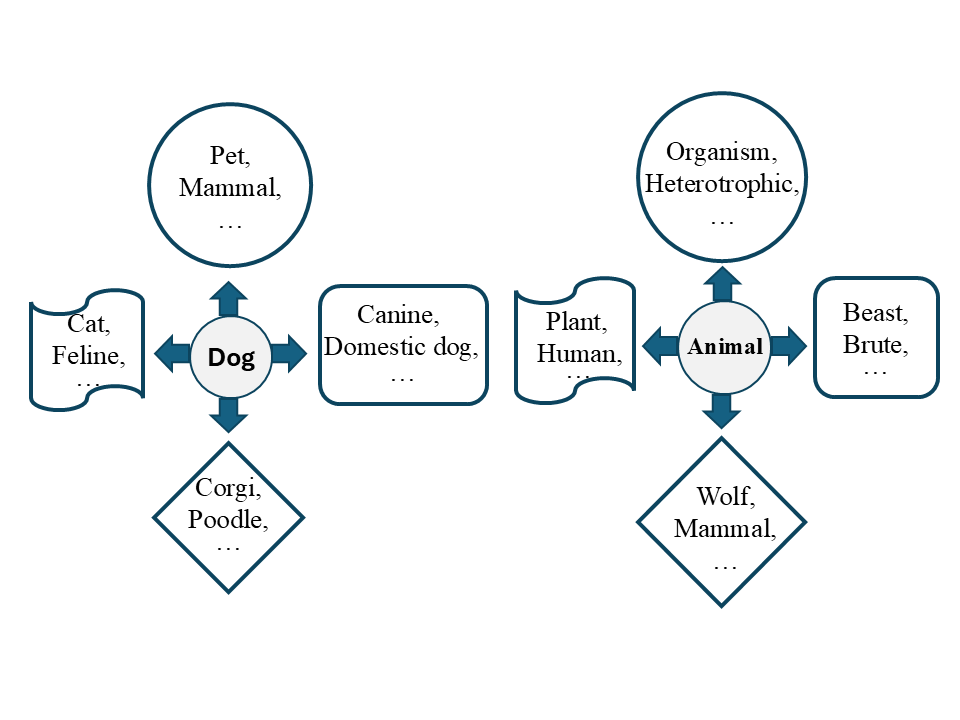}
\caption{}
\label{fig:fig5_relations_concepts}
\end{subfigure}
~
\begin{subfigure}[b]{0.9\linewidth}
\centering
\includegraphics[width=0.65\linewidth]{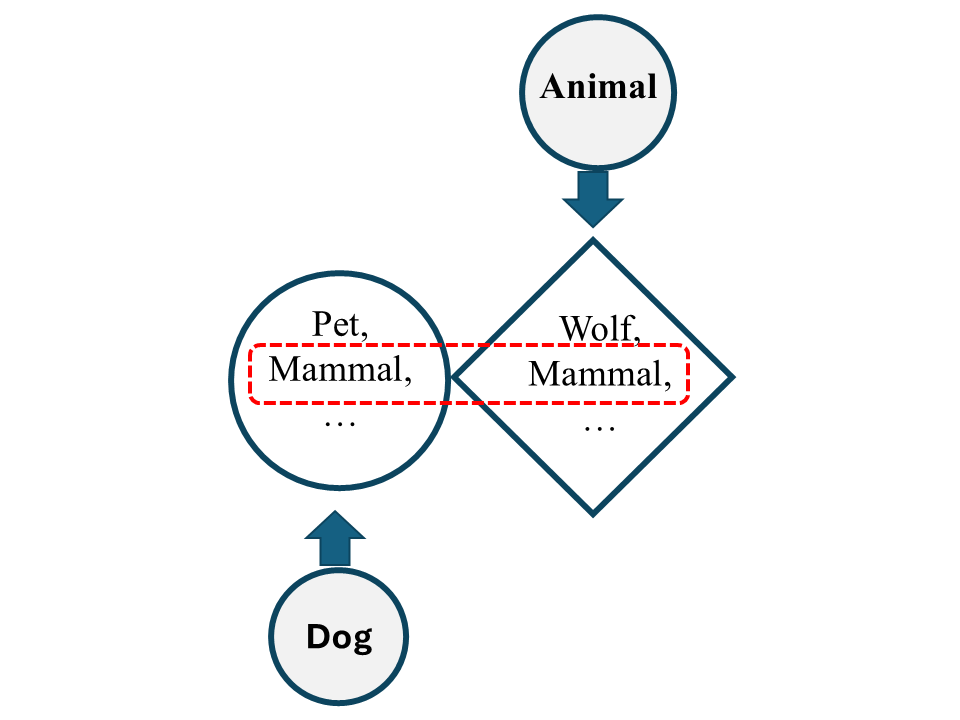}
\caption{}
\label{fig:fig5_intersections}
\end{subfigure}
\caption{Semantic relations according to the abstract categories of semantic compatibility and incompatibility. a) Semantic relations of entities of P and H and b) Intersection of sets of relations between entities of P and H maintaining hierarchy.}
\label{fig:fig5_intersec}
\end{figure}

For this, we need to develop the subgraphs of each entity of P and of H following the defined rules to find out if there are transitive connections that preserve SC or SI. The development of the subgraphs involves finding new concepts under the SC principles. These new concepts define sets of relations, or bags of relations, for each of the entities of P and H, as illustrated in Figure \ref{fig:fig5_relations_concepts}.

The idea is to check whether there are relationships connecting these entities through the transitivity property of general and concrete relationships, and the rules of equivalence and opposition that preserve SC and SI. In fact, there are four sets among which it is possible to search for intersections. If there is any intersection between these sets, it means that there are concepts that link entities from P and H, as illustrated in Figure \ref{fig:fig5_intersections}.

\begin{figure}[h!]%
\centering%
\includegraphics[width=\linewidth]{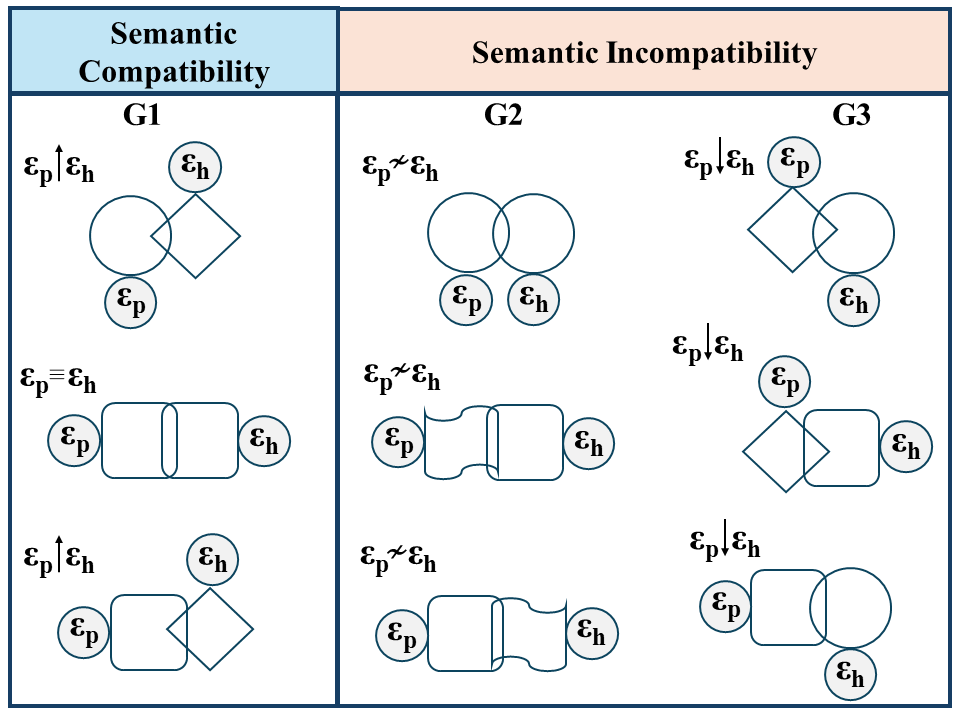}
\caption{Extension of relationships. $G_1$ captures equivalence relations (synonymy) and extends the generality of $\varepsilon_p$ using the concrete relationships of $\varepsilon_h$. $G_2$ captures co-hyponymy on the intersection of the general relations of both, and extended antonymy through equivalence relations. $G_3$ identifies more concrete relations from the concrete relations of $\varepsilon_p$ on general or equivalence relations of $\varepsilon_h$ and from equivalence relations of $\varepsilon_p$ on general relations of $\varepsilon_h$.
}\label{fig:fig6_ext_relations}
\end{figure}

However, not all intersections are valid, because the rules of transitivity must be respected for the SC to hold. Thus, of the 16 possible intersections between the 4 sets, only 9 are valid as shown in Figure \ref{fig:fig6_ext_relations}. Thus, new valid connections arise from these intersections with which it is possible to enrich the groups following their definition.

These intersections reveal deeper connections that inform our classification of entailment, allowing us to identify direct and indirect semantic relationships that might not be immediately apparent from ConceptNet alone and that might not be present in the LLM's conceptual network (Figure \ref{fig:fig6_ext_relations}).

Thus, the proposed groups define sets of semantic relationships that are abstracted into the concepts of SC and SI. We will refer to these groups as abstract groups, in the sense that they do not literally define the semantic relationships involved (e.g., we do not provide explicit information about what hyperonymy is), but rather abstract their compatibility by their association. Therefore, we are not seeking to say what these relationships are, but rather that there is a relationship at a higher level of abstraction that groups them together.%

\subsection{Calls to LLMs}

The framework allows us to address differentiated questions to an LLM by providing information from each of the categorized groups ($G_1-G_4$), ensuring independent multifaceted analysis. We apply prompting techniques using triplets from the groups we have defined, which has been shown to be the best way to provide information to these models \citep{DAI2025113060}. In the prompts, the definition of the groups and the list of triplets $G_1-G_4$ are included, see Figure \ref{fig:promts}.
The complete prompts can be found in \ref{app1:prompts}.

\begin{figure}[h!]
\centering

\begin{minipage}{13.5cm}
\fontfamily{courier}
\texttt{\scriptsize{You are an expert in Recognizing Textual Entailment over pairs of Premise and Hypothesis.
Based on the background information provide below, classify the relationship between the given Premise and Hypothesis as one of the following: "Entailment", "Neutral" or "Contradiction". Respond only using the template:
\{
"Answer":
\}.
Do not modify the template.}}

\texttt{\scriptsize{Premise and hypothesis to classify: }}

\texttt{\scriptsize{Premise:}} \scriptsize{\{$text_i$\}}

\texttt{\scriptsize{Hypothesis:}} \scriptsize{\{$hypothesis_i$\}}

\texttt{\scriptsize{Background Information:}} \scriptsize{\{$group\_definition_j$\}}

\texttt{\scriptsize{Word relations group:}} \scriptsize{\{$group\_relations_j$\}}
\end{minipage}

\caption{Prompt template for requests to LLMs. The variable \textit{i} runs through the number of examples in the datasets. The variable \textit{j} runs through the groups ($G_1 - G_4$) we propose.}
\label{fig:promts}
\end{figure}

\subsection{Inference in NLI}\label{sub:inference}
The LLM generates four different responses for each P-H pair, so a robust unification mechanism is essential. Our approach to consolidating the final responses of LLMs uses different techniques for final decision-making, evaluating the best reasoning process according to the abstract information provided.  To achieve this, we use three key strategies:
\begin{itemize}
    \item Majority Voting provides a robust and straightforward approach to selecting the most consistent answer.
\item Weighted Majority Voting (WMV) \citep{Dogan2019AWM} assigns dynamic weights to each classifier,
optimizing the final decision through ensemble learning.
\end{itemize}
\begin{itemize}
    \item Decision Tree Algorithm (Decision Tree) learns the rules, weighing up the correct lines of reasoning for the final answer.
\end{itemize}

For the latter two algorithms, additional sampling of pairs (P, H) is required for training only. The intuition behind this proposal is that each group ($G_1–G_4$) provides unique information about the relationships between P and H, resulting in different lines of reasoning. Combining these answers
maximizes the influence of lines of reasoning that lead to correct answers, resulting in a more reliable and informed final decision.

\section{Experimental results}\label{sec:exp}
We first describe the experimental setup used to assess our proposal and then present the results. The code used in this work is available at github \footnote{https://github.com/david-T-M/NLI\_with\_LLMs\_ConcepNet}.

\subsection{Experimental setup}
\subsubsection{LLMs}
\begin{figure}[h]%
\centering%
\includegraphics[scale=0.6]{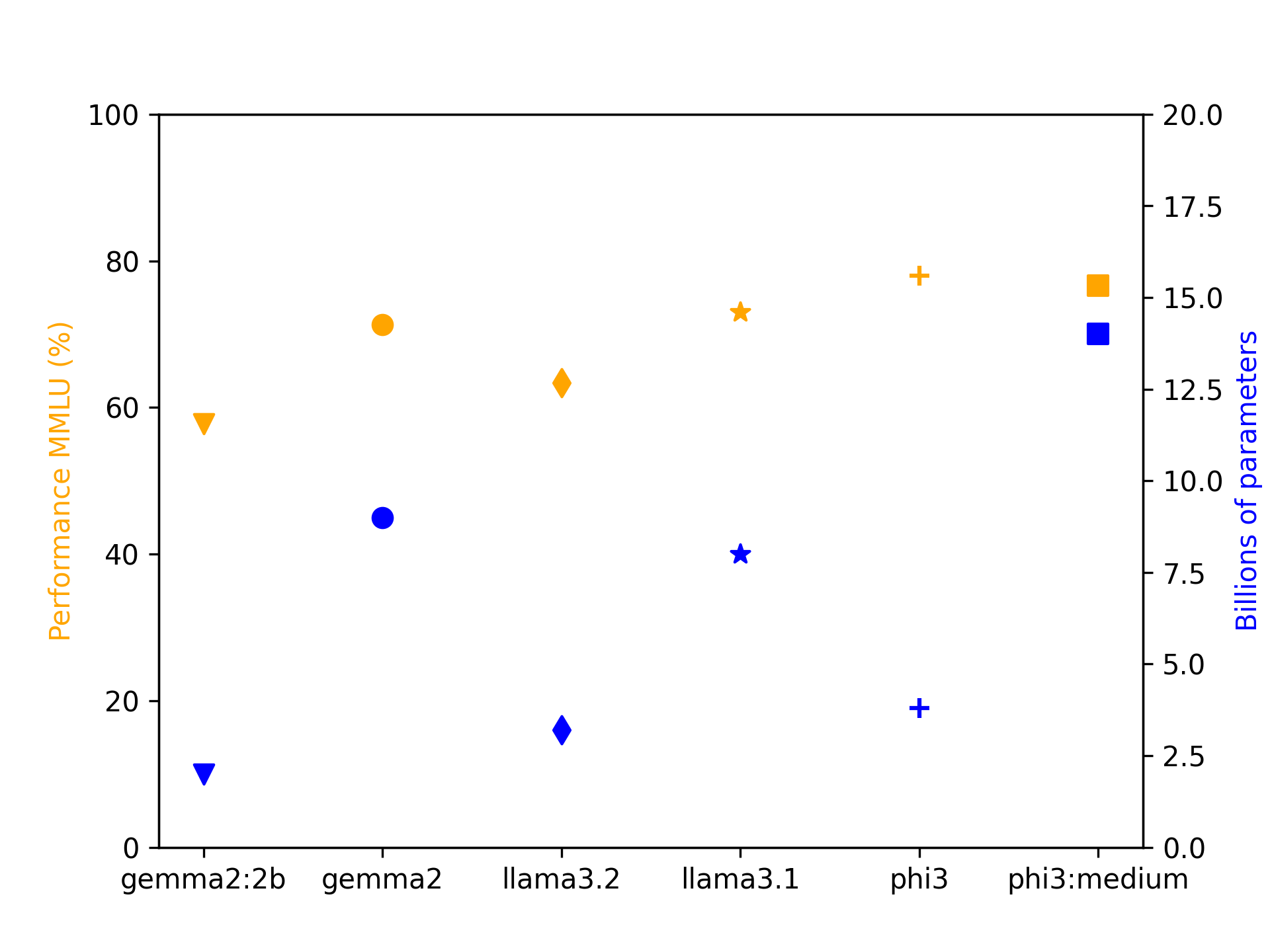}
\caption{Comparison of LLMs used in this study in terms of size and performance in MMLU.}\label{fig:fig8_MMLU}
\end{figure}

We selected LLMs from three families based on their size and performance on the \textbf{MMLU benchmark}\footnote{https://paperswithcode.com/sota/multi-task-language-understanding-on-mmlu}, which measures general knowledge in different subjects. These are: \texttt{gemma2:2b} and \texttt{gemma2} from Google, \texttt{llama3.2} and \texttt{llama3.1} from Meta%
, \texttt{phi3} and \texttt{phi3:medium} from Microsoft. Their corresponding size and performance on the MMLU benchmark are shown in Figure \ref{fig:fig8_MMLU}.

The \texttt{Ollama}\footnote{https://ollama.com} platform allows these models to be used directly, without additional configuration. All models ran on a Core i9 server with 128 GB of RAM and hosting an NVIDIA 4090 GPU.

\subsubsection{Datasets}
To assess NLI tasks, we selected the most widely used datasets: SICK \citep{marelli-etal-2014-semeval} and SNLI \citep{bowman-etal-2015-large}. These contain pairs of P-H with their respective labels: Entailment (E), Neutral (N), and Contradiction (C). We also chose datasets with two labels: entailment (E) and non-entailment (NE): SciTail \citep{Khot_Sabharwal_Clark_2018}, RTE \citep{wang-etal-2018-glue} and the SuperGLUE reference diagnostic dataset\footnote{https://super.gluebenchmark.com/diagnostics}.

With the exception of the diagnostic dataset, for which all example pairs were taken, random samples were generated, balancing the number of examples in each class, providing sufficient information to generate statistical tests, and reducing computational costs during processing. Table \ref{tab:samples} shows the datasets and their data.
\begin{table}[h]
\centering
\begin{tabular}{llll}
\hline
Dataset & Classes & Samples & Pairs for each sample \\ \hline
SNLI & 3 & 10 & 600; 200 per class \\ \hline
 SICK & 3 & 10 & 600; 200 per class \\ \hline
RTE & 2 & 10 & 400; 200 per class \\ \hline
Scitail & 2 & 10 & 400; 200 per class \\ \hline
Diagnostic & 2 & 1 & 1,104; 460 (E) y 644 (NE) \\ \hline
\end{tabular}
\caption{Datasets and samples for evaluating NLI}\label{tab:samples}
\end{table}

\subsection{Experimental analysis undertaken and results}\label{sec:resultados}

\subsubsection{Abstraction groups influence on LLMs}
The proposed abstraction groups were designed to capture semantic compatibility and incompatibility relationships in order to induce a line of reasoning in the LLMs. The hypothesis is that each abstraction group will induce responses with some tendency; for example, a tendency toward entailment or toward contradiction. Thus, the first question is to what extent these semantic abstraction groups influence the reasoning process of each model. For this analysis, LLMs were prompted using the template of Figure \ref{fig:promts} for each group.
\begin{figure*}[h!]%
\centering%
\includegraphics[width=\linewidth]{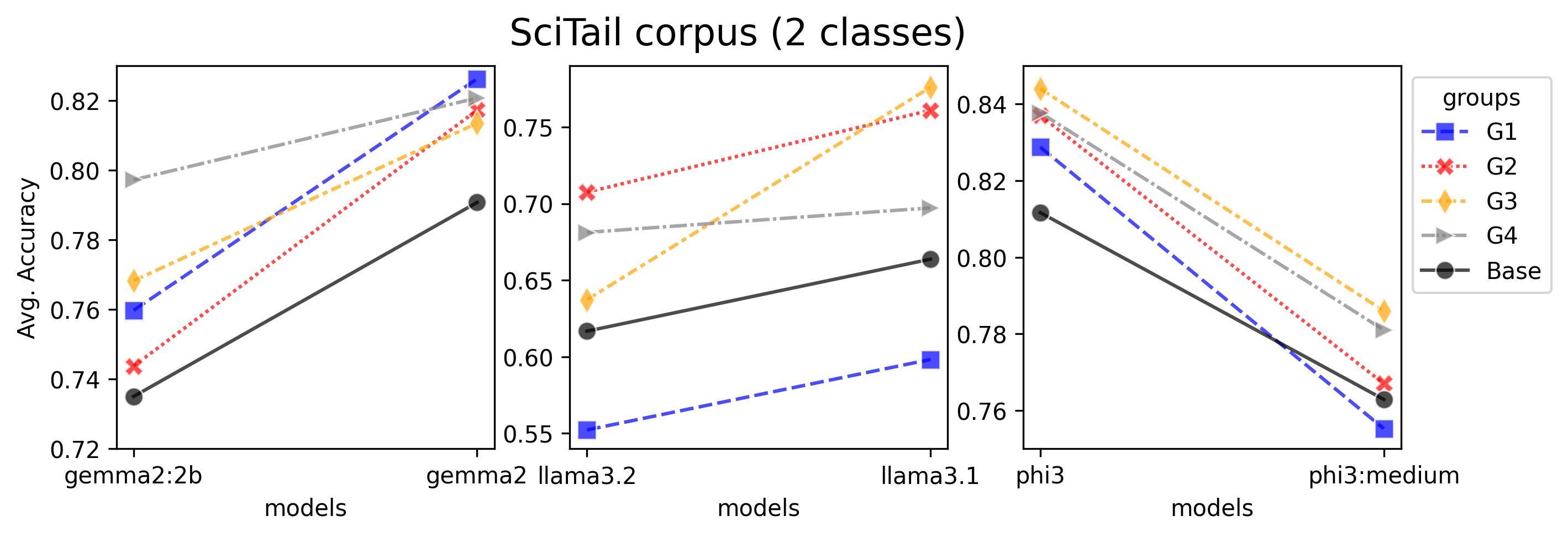}
\includegraphics[width=\linewidth]{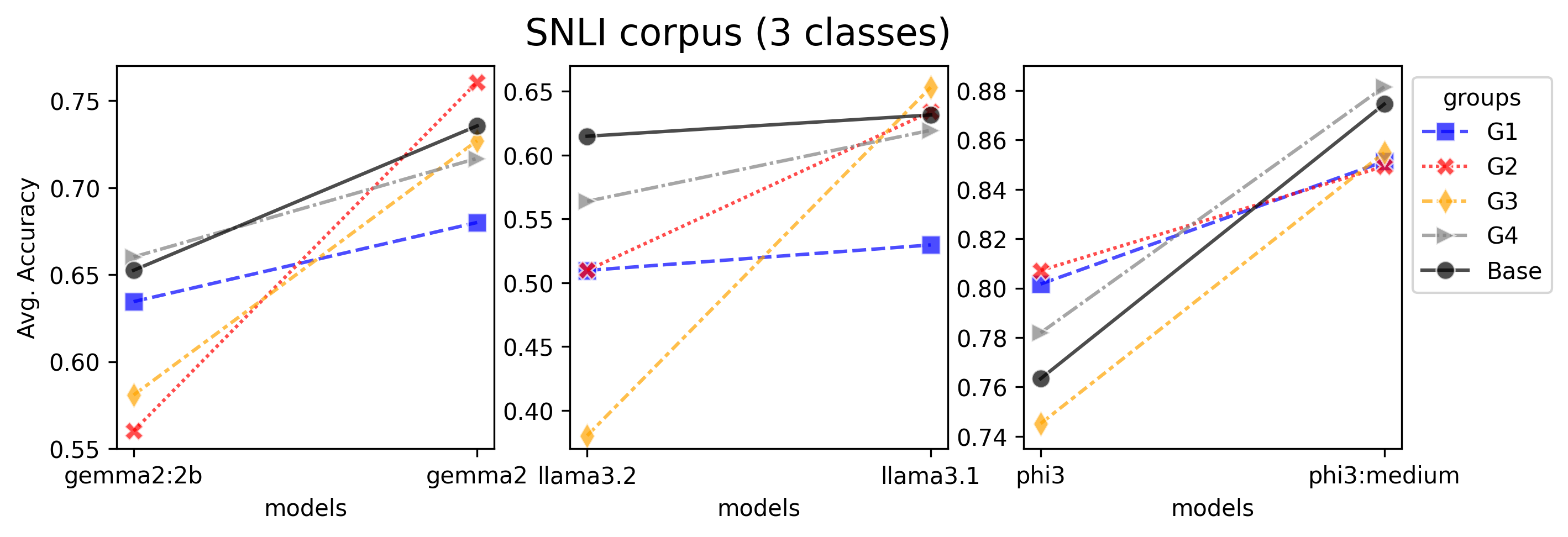}
\caption{Average accuracy of group influence on LLMs. Top row is the performance on the Scitail dataset by groups and baseline. Bottom row is the performance on the SNLI dataset by groups and baseline.}\label{fig:fig9_Perfgroups}
\end{figure*}

Figure \ref{fig:fig9_Perfgroups} summarizes the average performance of each LLM on different datasets under the influence of each abstraction group. The top row shows the average performance on the Scitail dataset (two labels), while the bottom row shows their performance on the SNLI dataset (three labels).
For the two-label dataset, higher accuracies are generally observed under the influence of abstraction groups and depending on the size of each model, except for the \texttt{Llama} models and the \texttt{phi3:medium} model with group $G_2$.
The situation changes slightly for the three-label dataset. In general, model performance drops compared to the baseline, except for some groups and some models.

\subsubsection{The GS\_DT system performance}
Figure \ref{fig:fig9_Perfgroups} shows that there always seems to be at least one group that performs better than the baseline on average. Thus, the second question is how to reach a consensus on the responses that each abstraction group elicits.

To unify the criteria, we experimented with the strategies proposed in Section \ref{sub:inference}. This allows us to deal with the variability of the LLMs' responses, when provided with information about the group, and identify the best strategy to decide on the correct answer.  In the case of WMV and DT, it is necessary to choose training and test samples from each dataset.
For majority voting, only the most frequent prediction is taken.

After several trials, we decided to use DT as the consensus mechanism, as it consistently obtained the best results.
Henceforth, we will refer to our proposal combining the abstraction groups and the decision tree as GS\_DT.

The performance of the GS\_DT system is shown for the three-label and two-label datasets in Tables \ref{corpus3labels} and \ref{corpus2labels}, respectively.

\begin{table}[h!]
\resizebox{13.5cm}{!}{
\begin{tabular}{lllllll}
\hline
            & \multicolumn{3}{c}{SNLI}              & \multicolumn{3}{c}{SICK}              \\ \hline
Models      & Baseline         & GS\_DT      & p-value & Baseline         & GS\_DT      & p-value \\ \hline
\texttt{gemma2}     & 73.6$\pm$1.7 & \textbf{78.9}$\pm$1.6 & $<$0.05  & 85.5$\pm$1.0 & \textbf{89.2}$\pm$0.9 & $<$0.001  \\
\texttt{gemma2:2b}   & 65.3$\pm$1.8 & \textbf{67.3}$\pm$1.7 & $<$0.001  & 76.5$\pm$1.0 & \textbf{84.3}$\pm$1.1 & $<$0.001  \\ \hline
\texttt{llama3.1}   & 63.1$\pm$1.6 & \textbf{74.7}$\pm$1.5 & $<$0.001  & 80.5$\pm$0.9 & \textbf{86.2}$\pm$0.6 & $<$0.001  \\
\texttt{llama3.2}    & 61.5$\pm$0.6 & \textbf{69.8}$\pm$1.2 & $<$0.001  & 67.0$\pm$0.8 & \textbf{78.2}$\pm$1.5 & $<$0.001  \\ \hline
\texttt{phi3:medium} & 87.4$\pm$1.3 & 87.8$\pm$1.5 & 0.5438  & 74.0$\pm$1.2 & \textbf{81.6}$\pm$1.4 & $<$0.001  \\
\texttt{phi3}        & 76.3$\pm$1.5 & \textbf{83.2}$\pm$2.1 & $<$0.01  & 83.4$\pm$2.0 & 85.0$\pm$1.3 & 0.0787  \\ \hline
\end{tabular}
}
\caption{Comparison of average performance between Baseline and GS\_DT for the datasets of 3 classes with p-values from Mann-Whitney statistical tests.}
\label{corpus3labels}
\end{table}

\begin{table}[h!]
\resizebox{13.5cm}{!}{
\begin{tabular}{lllllll}

\hline
 & \multicolumn{3}{c}{SciTail}     & \multicolumn{3}{c}{RTE}                        \\ \hline
Models      & Baseline         & GS\_DT      & p-value          & Baseline         & GS\_DT     & p-value          \\ \hline
\texttt{gemma2}      & 79.1$\pm$1.5 & \textbf{83.6}$\pm$1.2 & \textless{}0.001 & 88.1$\pm$1.2 & 88.4$\pm$1.2 & 0.8796  \\
\texttt{gemma2:2b}   & 73.5$\pm$1.2 & \textbf{79.8}$\pm$1.7 & \textless{}0.001 & 74.6$\pm$2.0 & 74.5$\pm$1.9 & 0.4956           \\ \hline
\texttt{llama3.1}   & 66.4$\pm$1.0 & \textbf{78.0}$\pm$1.7 & \textless{}0.001 & 76.2$\pm$2.0 & \textbf{81.3}$\pm$1.7 & \textless{}0.001 \\
\texttt{llama3.2}    & 61.7$\pm$1.7 & \textbf{71.2}$\pm$1.8 & \textless{}0.001 & 73.2$\pm$1.9 & \textbf{75.5}$\pm$1.9 & \textless{}0.05  \\ \hline
\texttt{phi3:medium} & 76.3$\pm$1.7 & \textbf{80.4}$\pm$1.5 & \textless{}0.001  & 87.5$\pm$1.1 & 88.3$\pm$1.5 & 0.3618           \\
\texttt{phi3}        & 81.2$\pm$1.1 & \textbf{84.2}$\pm$1.3 & \textless{}0.001 & 85.8$\pm$1.7 & 85.4$\pm$1.6 & 0.5699          \\ \hline
\end{tabular}
}\caption{Comparison of average performance between Baseline and GS\_DT for the datasets of 2 classes with p-values from Mann-Whitney statistical tests.}
\label{corpus2labels}
\end{table}

\subsubsection{Contribution to NLI tasks}
\begin{figure*}[h!]%
\centering%
\includegraphics[width=\linewidth]{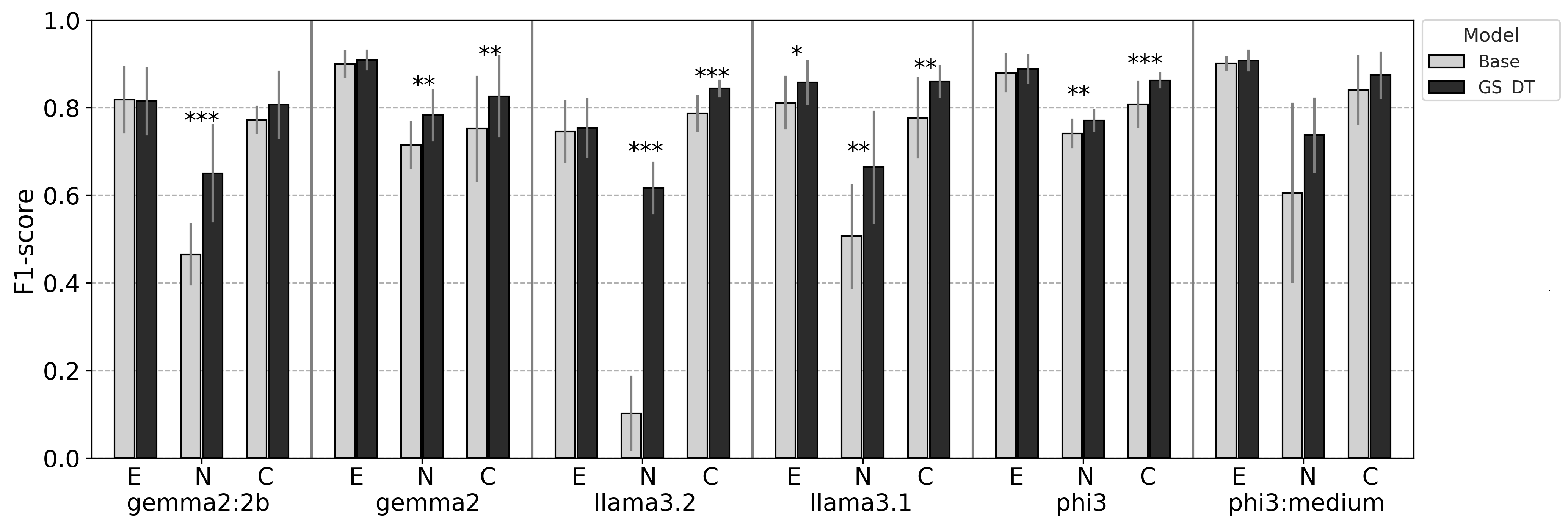}
\caption{Comparison of F1-score of classes in a 3-class datasets between Baseline and GS\_DT. $***$ indicates a p-value$<$0.001, $**$ a p-value$<$0.01, $*$ a p-value$<$0.05 obtained through the Mann-Whitney statistical test.}
\label{fig:fig10_f1_score_3corpus}
\end{figure*}

\begin{figure*}[h!]%
\centering%
\includegraphics[width=\linewidth]{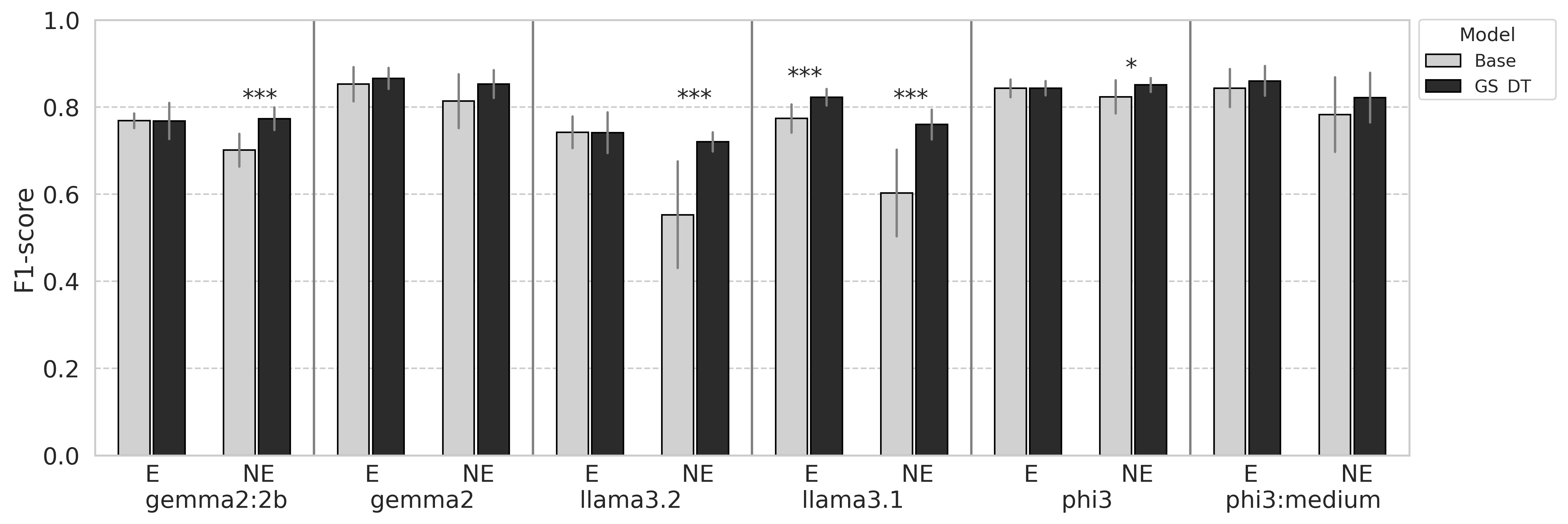}
\caption{Comparison of F1-score of classes in a 2-class datasets between Baseline and GS\_DT. $***$ indicates a p-value$<$0.001, $**$ a p-value$<$0.01, $*$ a p-value$<$0.05 obtained through the Mann-Whitney statistical test.}\label{fig:fig11_f1_score_2corpus}
\end{figure*}

We evaluate here the proposal's contribution to NLI tasks.
Figures \ref{fig:fig10_f1_score_3corpus} and \ref{fig:fig11_f1_score_2corpus} show the results of the F1-score metric by class. The results of the SNLI and SICK datasets are grouped together, and the results of the SciTail and RTE datasets are grouped together.

\subsubsection{Diagnostic dataset}
Another relevant question is the contribution of the system to linguistic aspects using the diagnostic dataset. To do this, we analyzed the impact of the system on the linguistic categories of the dataset and its contribution to improvement and loss with respect to the baseline.

\begin{figure}[h]%
\centering%
\includegraphics[width=\linewidth]{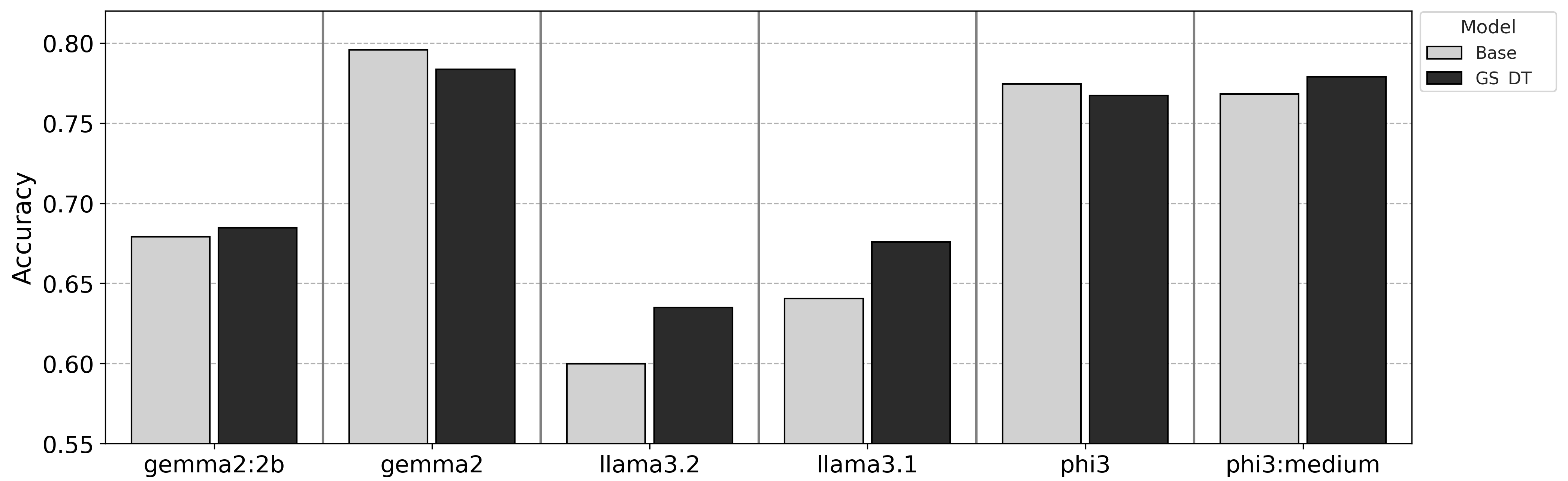}
\caption{Accuracy of LLMs on diagnostic dataset}\label{fig:fig12_Perf_diag}
\end{figure}

The diagnostic dataset has 1104 P,H pairs. The Entailment class has 460 examples and the Not-Entailment class has 644 examples.
In Figure \ref{fig:fig12_Perf_diag}, we show the accuracy of our proposal against the baseline.

Figure \ref{fig:fig12_Perf_diag} shows that our proposal helps in some cases and is detrimental in others. Therefore, we investigated how much improvement or loss there is by comparing the examples that the LLM baseline answers correctly and incorrectly against our proposal. Table \ref{table:improve_confuse} shows the results.

\begin{table}[h!]
\resizebox{13.5cm}{!}{
\begin{tabular}{lllllll}
\hline
        & \texttt{gemma2:2b} & \texttt{gemma} & \texttt{llama3.2} & \texttt{llama3.1} & \texttt{phi3} & \texttt{phi3:medium} \\
        \hline
Improve & 17.4\%     & 20.2\%  & 26.2\%     & 18.9\%     & 13.3\% & 14.5\%        \\
\hline
Loss    & 6.0\%       & 6.5\%   & 11.5\%     & 5.1\%      & 4.8\%  & 3.0\%     \\
\hline
\end{tabular}}\caption{Overall improvement and loss for the diagnostic dataset. Improvement: refers to the proportion of correct predictions under the GS\_DT proposal out of the total number of failures of the baseline model. Loss: refers to the proportion of failures under the GS\_DT proposal out of the total number of correct predictions of the baseline model.}
\label{table:improve_confuse}
\end{table}

\subsubsection{Ablation study}
Finally, an ablation study was conducted to identify the contributions of the proposal compared to a vanilla and the state-of-the-art models on the diagnostic dataset. We also analyzed the impact of our abstract groups against direct relations in ConceptNet.

Table \ref{comp_ll32_phi3_m} shows the performance obtained with the \texttt{llama3.2} and \texttt{phi3: medium} models. We added the model \texttt{Vega v2} (leader in the RTE benchmark), and a model \texttt{Majority class} to compare with all Direct Relations (DR) in ConcepNet, all relations from Abstract Groups gathered together (AR), and our proposal (GS\_DT).
\begin{table}[H]
\centering
\begin{tabular}{llll}
\hline
Models                                         & Accuracy       & \begin{tabular}[c]{@{}l@{}}Entailment\\ F1-score\end{tabular} & \begin{tabular}[c]{@{}l@{}}Not\_entailment\\ F1-score\end{tabular} \\ \hline
\texttt{Vega v2}  & 43.29   & 56.76   & 17.63                                                              \\
\texttt{Majority class}                        & 58.33          & 0                                                         &   73.68                                                                \\ \hline
\texttt{llama3.2} baseline    & 59.98          & 64.40                                                         & 54.30                                                              \\
+Direct relationships (DR)                          & 59.78          & 64.53                                                         & 53.55                                                              \\
+Abstract relations (AR)                             & 61.50          & \textbf{65.24}                                                & 56.85                                                              \\
+GS\_DT                                             & \textbf{63.49} & 62.92                                                         & \textbf{64.05}                                                     \\ \hline
\texttt{phi3:medium} baseline & 76.81          & 77.26                                                         & 76.34                                                              \\
+Direct relationships   (DR)                        & 77.26          & \textbf{78.07}                                                & 76.38                                                              \\
+Abstract relations  (AR)                           & 77.35          & 77.95                                                         & 76.72                                                              \\
+GS\_DT                                             & \textbf{77.89} & 77.97                                                         & \textbf{77.81}                                                     \\ \hline
\end{tabular}\caption{Ablation results on diagnostic dataset. }
\label{comp_ll32_phi3_m}
\end{table}

\section{Discussion}\label{sec:disc}

The implementation of our approach combines prompting techniques with consensus mechanisms.
In line with \citep{li-etal-2023-making, kasner-etal-2023-mind, huang-etal-2024-enhancing, mu-etal-2024-ddprompt} methodology, we explore different lines of reasoning for the same question. In our work, we construct abstract groups of semantic relationships that influence the reasoning of LLMs.

\subsection{Consensus mechanism}

\begin{figure}[h!]%
\centering%
\includegraphics[width=\linewidth]{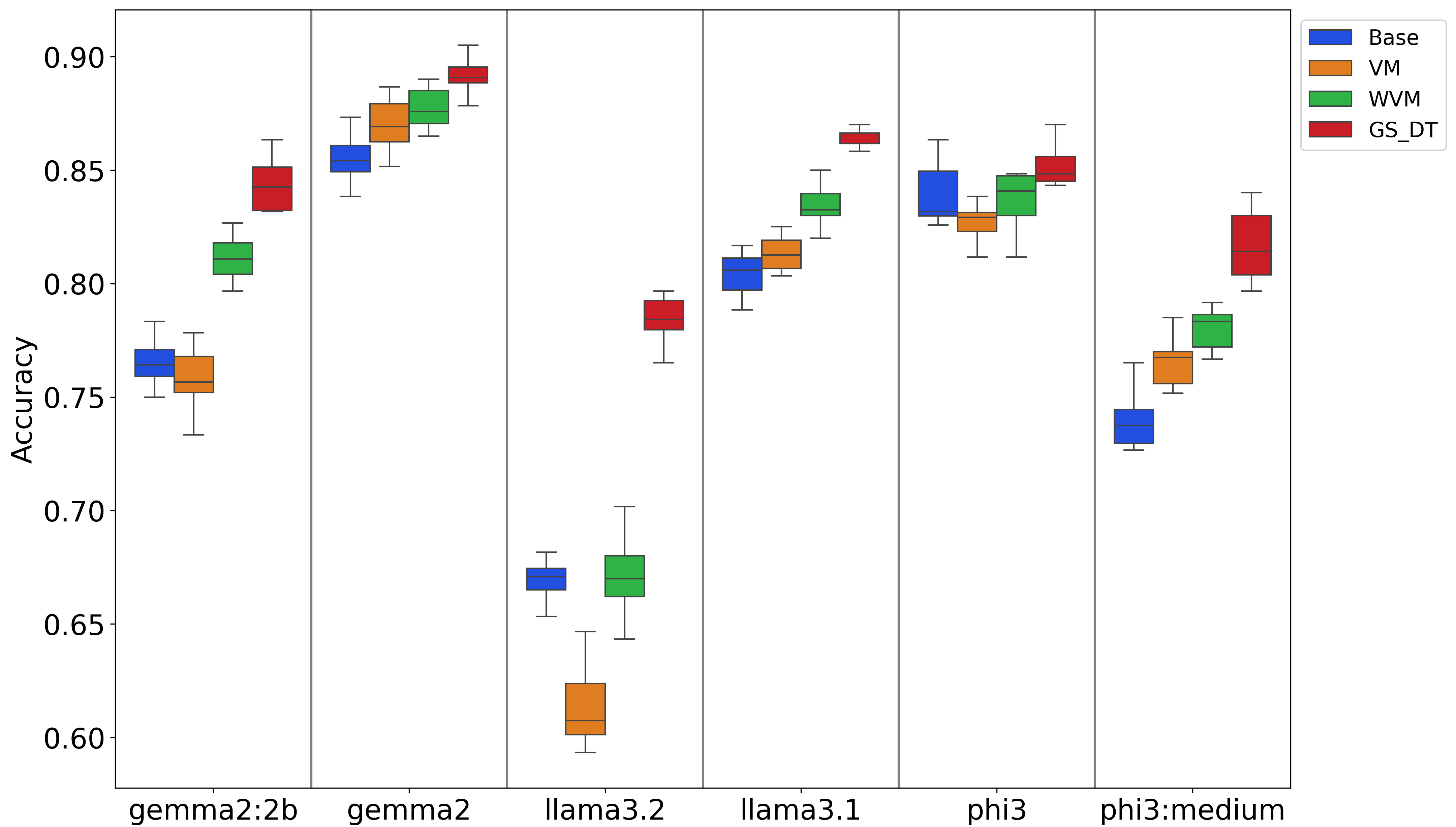}
\caption{Comparison of consensus mechanism performance: VM vs WVM vs DT - dataset SICK}\label{fig:fig13_p_SICK}
\end{figure}

The choice for the consensus mechanism was made after several trials that consistently showed that DT was the best option. Figure \ref{fig:fig13_p_SICK} illustrates the point for the SICK dataset.

Figure \ref{fig:fig13_p_SICK} shows how a majority voting strategy is not an adequate mechanism for making an informed decision. In a traditional majority voting scheme, the assumption is that when asked the same question, regardless of how the model is prompted, it should arrive at the same answer and be consistent in its reasoning. In other words, the model would have to follow the same line of reasoning regardless of how it is prompted.

What our results show is that, when asked the same question, at least one of the groups provides relevant disjunctive information about the elements of the question (Figure \ref{fig:fig9_Perfgroups}). Therefore, the models follow different lines of reasoning in each instance. However, what Figure \ref{fig:fig13_p_SICK} shows is that if the decision is made based on the majority of votes, the result is practically the same or worse than the baseline vote: majority voting fails, in contrast to \citep{xue-etal-2023-dynamic,wang2023selfconsistencyimproveschainthought,
wang-etal-2024-soft}. The weighted majority voting scheme helps a little. However, it is unable to resolve ambiguous situations, and the weighting pattern is clearly inconsistent across the models.

What the decision tree does is identify the pattern of agreement between groups that leads to the correct decision. If the abstraction groups were not consistently providing useful information, there would be no pattern to identify. Therefore, abstraction groups fill knowledge gaps in the models, prompting them to take consistent lines of reasoning and give correct answers consistently.

A final consideration regarding the relevance of using the decision tree for final decision-making in an LLM regards recent strategies based on debate. A strategy to improve LLMs' responses is carried out through multi-agent discussion \citep{wang-etal-2024-rethinking-bounds}, based on majority voting. In this sense, our consensus mechanism uses an explicit hierarchical decision tree to avoid the inconsistencies of debate among LLMs, such as error propagation or judge bias \citep{wang-etal-2024-rethinking-bounds}, and to ensure greater control and reliability in responses. Thus, we focus on compensating for the inherent shortcomings of LLMs, regardless of their size: the agent becomes stronger, which could lead to better arguments in scenarios involving debate between agents. This allows us to achieve robust results without incurring the high costs associated with use of industry standard (larger) LLMs.

\subsection{Performance}

Simply incorporating direct information from external resources is not enough (Table \ref{comp_ll32_phi3_m}); a strategy is also needed to address the shortcomings of LLMs. DT\_GS in our framework acts as a meta-reasoner that prioritizes high-confidence routes of reasoning and assigns greater weight to relationship groups where there is strong consensus. Furthermore, it learns trade-offs, e.g., when multiple LLM decisions conflict, the tree learns to combine signals from these decisions about the groups.

The results (Tables \ref{corpus3labels} and \ref{corpus2labels}) show that \texttt{llama3} models make better use of the new information in their reasoning processes.
With DT\_GS, their results range from an improvement of \textbf{5.7} to \textbf{11.6} p.p. in the \textit{SICK}, \textit{SNLI} and \textit{SciTail} datasets compared to baseline. We were able to outperform the baseline model in these datasets, but for the RTE dataset, even if higher, results are not significant, except for the \texttt{llama3} models.

Overall, GS\_DT is an effective strategy for improving the performance of language models, particularly those with a lower initial performance. In models that are already optimized, however, the impact is marginal; a conjecture is that this may be due to training data contamination.

Also, GS\_DT is the most balanced and robust proposal, providing the best overall performance in terms of accuracy and distinguishing cases where there is no entailment relationship (Figures \ref{fig:fig10_f1_score_3corpus} and \ref{fig:fig11_f1_score_2corpus}).

Finally, the SuperGLUE benchmark \citep{wang2019} provides a diagnostic dataset specifically designed to assess linguistic competence in NLI.
The analysis reveals that not only are there profound variations in the performance of models on the diagnostic dataset, but they also have profound limitations in semantic understanding (Table \ref{comp_ll32_phi3_m}).
GS\_DT improves and balances the inference in the diagnostic dataset (Table \ref{table:improve_confuse}); thereby better benefiting the not entailment class.

\subsection{Impact of abstraction}
Current evaluation methods for LLMs primarily assess inferential capabilities through outcome-based metrics. However, this fails to capture the complexity of their underlying reasoning processes. Notably, when faced with straightforward questions that require the application of general knowledge, these models often struggle to abstract and utilize relevant information, exposing a significant shortcoming in their abstract reasoning abilities \citep{NEURIPS2024_da5498f8,Lee_2025_abstraction}.

Our approach aims at enriching LLMs with relevant, abstract and structured knowledge to improve their performance in NLI tasks.  Abstraction not only improves immediate performance; it redefines how LLMs integrate knowledge and how they use it. In contrast to techniques such as Abstraction-of-Thought (AoT) \citep{hong-etal-2024-abstraction}, which relies on the model generating its own abstractions (with the risk of inconsistency or hallucinations), our proposal of external abstract categories of semantic relations structures the reasoning space and allow us to trace the decision-making process, something impossible with raw data.

When working with raw relationships, LLMs tend to get stuck in literal associations and fail to capture the semantic depth \citep{KIM2024112608}.
In contrast, by structuring knowledge into abstract categories that mimic human reasoning, models benefit from new concepts or connections between concepts and avoid using learned lexical patterns; in other words, providing novel relationships could prevent a tendency toward overfitting or shortcuts based on certain clues in the premise or hypothesis.

Similarly, abstractions resolve ambiguities by identifying that two concepts can be related in multiple ways, and even detect contradictions and neutrality with greater precision (Figures \ref{fig:fig10_f1_score_3corpus} and \ref{fig:fig11_f1_score_2corpus}), areas where LLMs often fail.

The results (Table \ref{comp_ll32_phi3_m}) demonstrate that GS\_DT is a robust approach
as it balances between generalization and specificity, minimizing false positives.
The AR approach, on the other hand, is superior in tasks where the accurate identification of \textit{entailment} is crucial. Both approaches, GS\_DT and AR, outperform DR and the Baseline consistently, demonstrating the significant impact of their structural improvements.
The latter approaches are limited by their conceptual rigidity and reduced adaptability, as they are based on static, predefined relationships.

\section{Conclusions}\label{sec:concl}
NLI tasks require analyzing complex implicit/explicit relationships, demanding both the LLM's internal knowledge and external commonsense \citep{liu-etal-2021-commonsense}. Our strategy of abstracting categories to guide LLM decisions, improves contradiction detection and entailment classification.
Crucially, LLMs need structured knowledge and not merely more data to bridge reasoning gaps. Our abstraction framework convert raw text into interpretable patterns, correcting biases and boosting generalization. While LLMs’ pretrained knowledge is insufficient for NLI, external data alone also falls short. Our hybrid approach directs attention to overlooked word relationships, enabling models to synthesize missing information.
The quality of external knowledge is relevant because there are biases of incomplete or culturally biased relationships, which affects the generalization of abstract categories. Similarly, its coverage is limited and may lack well-defined relationships, reducing the effectiveness of the method.
There is also the problem of ambiguity in conceptual boundaries, i.e., some relationships may fit into several categories \citep{ILIEVSKI2021107347}.

The results reveal the benefits of our semantic relation abstraction framework for compensating for knowledge gaps in LLMs. The future lies not in expanding model size \citep{li-etal-2022-systematic}, but in creating a semantic scaffolding that mimics the flexibility of human thinking. We hope that our proposal will enable the development of more robust agents and interpretable reasoning, guiding AI toward a reliable understanding of language.

\newpage
\bibliographystyle{elsarticle-num}
\bibliography{revibiblio}

\newpage
\appendix
\section{Prompts}
\label{app1:prompts}

\begin{figure}[h]%
\centering%
\includegraphics[scale=0.35]{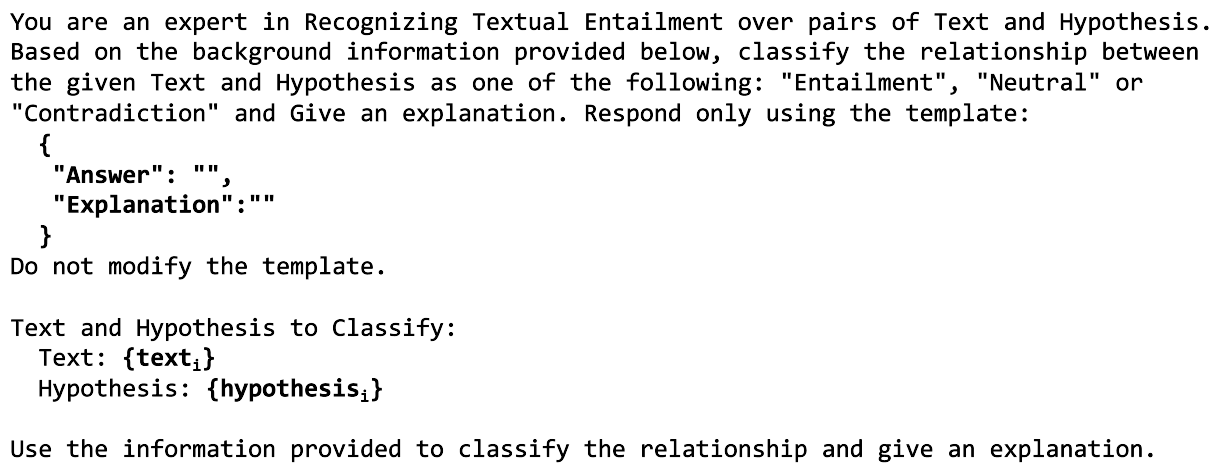}
\caption{Prompt Baseline}
\label{fig12}
\end{figure}

\begin{figure}[h]%
\centering%
\includegraphics[scale=0.35]{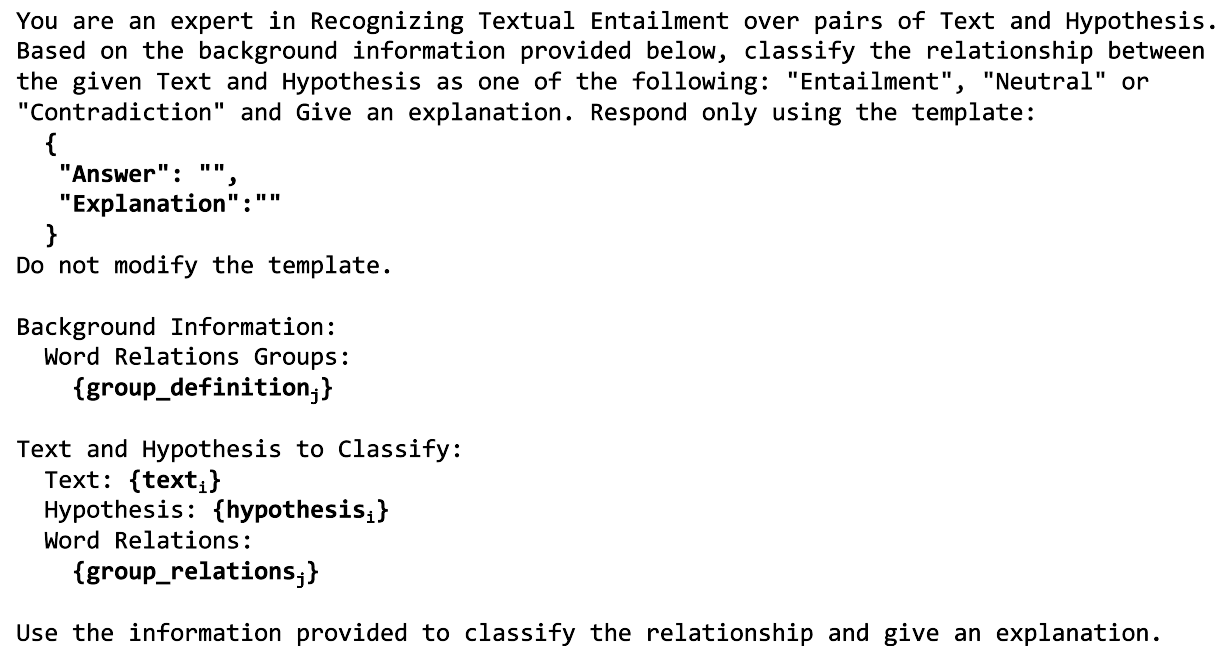}
\caption{Prompt with external information }\label{fig13}
\end{figure}

\begin{figure}[h]%
\centering%
\includegraphics[scale=0.4]{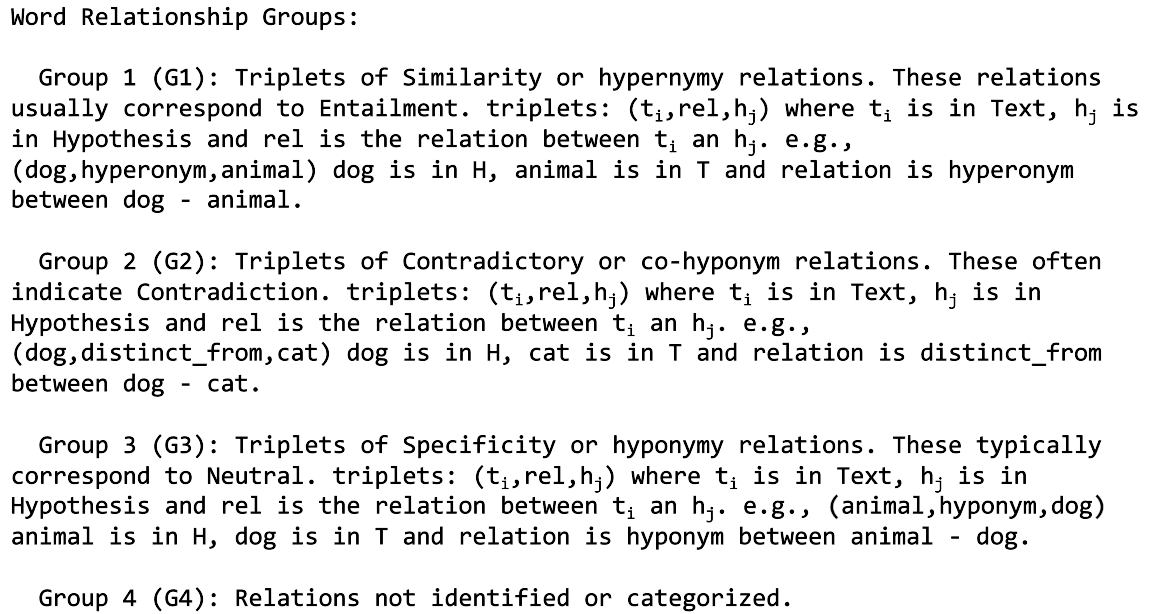}
\caption{Definitions of the groups}\label{fig14}
\end{figure}

\end{document}